\pdfoutput=1

\documentclass[11pt]{article}

\usepackage[final]{acl}

\usepackage{times}
\usepackage{latexsym}

\usepackage[T1]{fontenc}

\usepackage[utf8]{inputenc}

\usepackage{microtype}

\usepackage{inconsolata}

\usepackage{graphicx}

\usepackage{amsmath}
\usepackage{amsthm}
\usepackage{booktabs}
\usepackage{algorithm}
\usepackage{algorithmic} 
\usepackage{multirow}
\usepackage{amssymb}
\usepackage{xcolor} 
\usepackage{soul} 
\usepackage{colortbl} 
\usepackage{array}    
\usepackage{url}      

\title{A Symbolic Adversarial Learning Framework for\\ Evolving Fake News Generation and Detection\thanks{Accepted to EMNLP 2025 Main Conference.}}

\author{
  Chong Tian \\
  MBZUAI \\
  Abu Dhabi, UAE \\
  \texttt{Chong.Tian@mbzuai.ac.ae} \\\And
  Qirong Ho\thanks{~Corresponding Authors.} \\
  MBZUAI \\
  Abu Dhabi, UAE \\
  \texttt{Qirong.Ho@mbzuai.ac.ae} \\\And
  Xiuying Chen\footnotemark[2] \\
  MBZUAI \\
  Abu Dhabi, UAE \\
  \texttt{Xiuying.Chen@mbzuai.ac.ae} \\
}

\definecolor{lightblue}{RGB}{173,216,230}
\definecolor{lightred}{RGB}{255,200,200}
\newcommand{\hlb}[1]{\sethlcolor{lightblue}\hl{#1}}  
\newcommand{\hlr}[1]{\sethlcolor{lightred}\hl{#1}}   

\begin{document}
\maketitle
\begin{abstract}
Rapid LLM advancements heighten fake news risks by enabling the automatic generation of increasingly sophisticated misinformation. Previous detection methods, including fine-tuned small models or LLM-based detectors, often struggle with its dynamically evolving nature. In this work, we propose a novel framework called the \textit{Symbolic Adversarial Learning Framework (SALF)}, which implements an adversarial training paradigm by an agent symbolic learning optimization process, rather than relying on numerical updates. SALF introduces a paradigm where the generation agent crafts deceptive narratives, and the detection agent uses structured debates to identify logical and factual flaws for detection, and they iteratively refine themselves through such adversarial interactions. Unlike traditional neural updates, we represent agents using agent symbolic learning, where learnable weights are defined by agent prompts, and simulate back-propagation and gradient descent by operating on natural language representations of weights, loss, and gradients. Experiments on two multilingual benchmark datasets demonstrate SALF's effectiveness, showing it generates sophisticated fake news that degrades state-of-the-art detection performance by up to 53.4\% in Chinese and 34.2\% in English on average. SALF also refines detectors, improving detection of refined content by up to 7.7\%. We hope our work inspires further exploration into more robust, adaptable fake news detection systems.
\end{abstract}

\section{Introduction}
\label{sec:introduction}

\begin{figure}
    \centering
    \includegraphics[width=1\columnwidth]{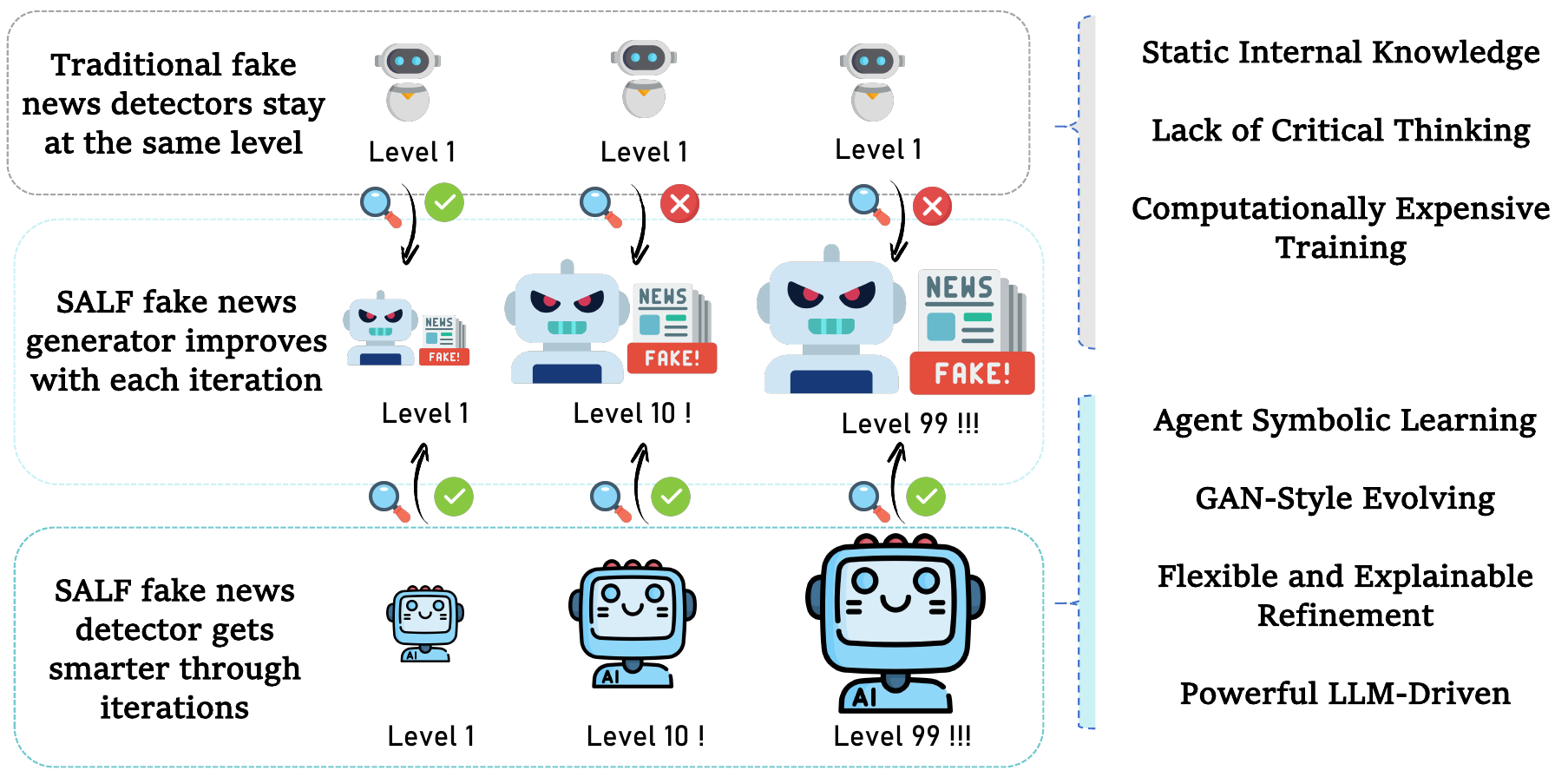} 
    \caption{While existing fake news detectors remain static and fail to keep up with the increasingly sophisticated fake news, our SALF framework showcases continuous and effective evolution. 
    }
    \label{fig:SALF demo}
\end{figure}

The dissemination of fake news, defined as fabricated information mimicking legitimate news, has become an increasingly pervasive issue, particularly with the rise of social media as a primary source of information. 
Its far-reaching consequences extend to influencing elections~\cite{allcott2017social}, public health~\cite{naeem2021exploration}, and economic stability~\cite{mwangi2023technology}.
Worse even, the rise of LLMs dramatically lowers the barriers to generating sophisticated fake news~\cite{sun2024exploring}, and the fake news has evolved to be more deceptive~\cite{sciannamea2020fake,liu2024tiny,liu2024skepticism}.

Fighting against fake news has garnered significant attention in recent years~\cite{zhou2020survey,kumar2018false,chen2023spread}, and existing approaches can generally be classified into two categories.
One paradigm employs smaller models specifically fine-tuned for the fake news detection task~\cite{hu2024bad,aggarwal2020classification}, while the other focuses on designing more effective prompts for LLMs~\cite{su2023fake,hu2024bad}.
However, these methods often struggle to efficiently combat the evolving nature of fake news~\cite{guo2021does}. 
Smaller language models are typically trained on corpora collected during a specific period, limiting their ability to generalize to new fake news or unseen data~\cite{o2018language}. 
Similarly, for LLMs, even carefully crafted prompts designed for detecting fake news in a specific context may fail to adapt effectively to fake news or misinformation generated in different temporal or thematic~contexts.

Hence, in this work, we propose a Symbolic Adversarial Learning Framework (SALF), consisting of a fake news generation agent and a detection agent, aimed at addressing the above challenges.
Both agents are LLM-based, leveraging the strong semantic understanding capabilities of these models.
As the name suggests, our framework incorporates an adversarial concept similar to GANs~\cite{goodfellow2014generative}, where the generation agent crafts deceptive narratives, and the detection agent engages in identifying logical flaws and inaccuracies.
In this setup, both agents undergo continuous improvement through adversarial interactions.
However, unlike traditional GANs, where the update process relies on numerical neural network computations, updating LLMs directly through such methods is computationally expensive and impractical. 
To overcome this limitation, we extend the agent symbolic learning work~\cite{zhou2024symbolic} to our adversarial learning framework, where the learnable weights are defined as agent strategies, represented by prompts in this work.
Agent symbolic learning simulates backpropagation and gradient descent by operating on natural language representations of weights, losses, and gradients. 
In other words, the adversarial training process is achieved by iteratively refining the prompts for both the generation and detection agents based on their performance.
This enables the generation agent to craft increasingly deceptive narratives, while the detection agent enhances its ability to identify logical inconsistencies and inaccuracies through structured debates. 
This symbolic approach allows for a more interpretable and adaptive adversarial training process.

The key contributions of this work are as follows: 
First, we innovate to extend a recently proposed Agent Symbolic Learning framework to a GAN-like adversarial training paradigm, creating the Symbolic Adversarial Learning Framework (SALF) and proving its feasibility and effectiveness. Second, we apply SALF to fake news detection and generation, where it improves through interactions between a fake news generator and detector, adapting to the evolving nature of fake news and contributing to overcoming the limitations of other static models. Finally, we implemented comprehensive experiments to prove the effectiveness of the SALF framework. To be specific, the SALF generator generates sophisticated fake news that degrades state-of-art detection performance by up to 53.4\% on the Chinese dataset and 34.2\% on the English dataset, on average, while the SALF refined generator has a 7.7\% detection improvement towards these refined fake news.

\section{Related Work}
\label{sec:related_work}

\subsection{Fake News Detection}
Early fake news detection methods primarily relied on handcrafted linguistic features combined with classic machine learning classifiers~\cite{qian2018neural,yu2017convolutional}. These approaches captured surface-level cues, such as specific word usage or sentence structures, and achieved promising results in controlled scenarios. However, their performance often deteriorated when applied to unstructured social media data or adversarially crafted misinformation~\cite{dsouza2022social,bhatt2022fake}.
Subsequent research introduced smaller language models with enhanced reasoning capabilities~\cite{jin2022finegrained,zhu2022generalizing}, which allowed for the detection of more subtle logical inconsistencies within textual content. Additionally, efforts to integrate multimodal data, such as images and source metadata, further improved the robustness of fake news detection systems~\cite{zheng2022multimodal}.
More recently, the strong semantic understanding capabilities of LLMs have been leveraged for fake news detection. For example, the work~\cite{ma2024fake} utilized LLMs to analyze contextual relationships and detect nuanced misinformation. 
However, such methods rely on static prompts and the inherent knowledge of specific LLMs, limiting their ability to adapt and improve through self-learning and constraining their performance in evolving misinformation scenarios.
The work most related to ours is ~\cite{wang2024llm}, which proposed LLM-GAN, an iterative framework that adversarially optimizes both the fake news generator and detector.
However, LLM-GAN uses direct fake news detection without critical thinking, limiting its adversarial optimization due to potential inherent biases and knowledge boundaries of specific LLMs.
Moreover, it focuses solely on enhancing detector performance while neglecting to evaluate the generator component, resulting in a partial detection method that inadequately adapts to evolving fake news.

\subsection{Fake News Generation}
As a countermeasure to fake news detection, research on fake news generation has emerged, serving as a critical tool for benchmarking and improving detection models. 
These works simulate the strategies employed in real-world misinformation campaigns, enabling researchers to test and enhance the robustness of detection systems against evolving and sophisticated fake news~\cite{wanda2024deepnews,wang2024style}.
Early approaches to fake news generation relied on template-based or rule-based methods~\cite{shu2021fact}, producing fabricated content with limited diversity and realism. 
With advancements in natural language processing, modern fake news generation techniques have adopted generative models, such as GPT-series LLMs, capable of crafting highly sophisticated and contextually coherent misinformation~\cite{huang2024fakegpt,pan2023risk}.
For instance, the study~\cite{huang2024fakegpt} demonstrates ChatGPT's proficiency in generating high-quality fake news samples through various prompting methods. Other recent efforts also leverage LLMs to generate adversarial examples through stylistic attacks, primarily to enhance detector robustness~\cite{wu2024fake, park2025adversarial}. 


However, these methods lack adaptability: their predefined strategies, such as carefully crafted LLM prompts, fail to emulate the dynamic nature of real-world fake news, resulting in generated content that is superficial and relatively crude.

\subsection{Automatic Prompt Engineering}
Prompt engineering has become a pivotal technique for enhancing the performance of LLMs across diverse tasks. 
Traditionally, this process involves manually crafting prompts to elicit desired behaviors, which is both time-consuming and reliant on human expertise~\cite{giray2023prompt}. 
To address these limitations, recent research has focused on automating the prompt engineering process through innovative methods.
One such method is the Automatic Prompt Engineer~\cite{chen2024reprompt}, which leverages LLMs to autonomously generate and refine prompts. 
Similarly, RePrompt~\cite{chen2024reprompt} introduces a novel approach for optimizing prompts, enabling LLMs to learn domain-specific strategies for tasks like PDDL generation~\cite{guan2023leveraging} and travel planning.
Extending this, agent symbolic learning~\cite{zhou2024symbolic} treats prompts as learnable components, enabling agents to dynamically adjust their prompts and configurations, thereby enhancing adaptability to new tasks. 
In this work, we introduce the automatic agent symbolic learning process into adversarial fake news generation and detection settings.



\section{Methodology}
\label{sec:methodology}

\begin{figure*}[ht]
    \centering
    \includegraphics[width=\textwidth]{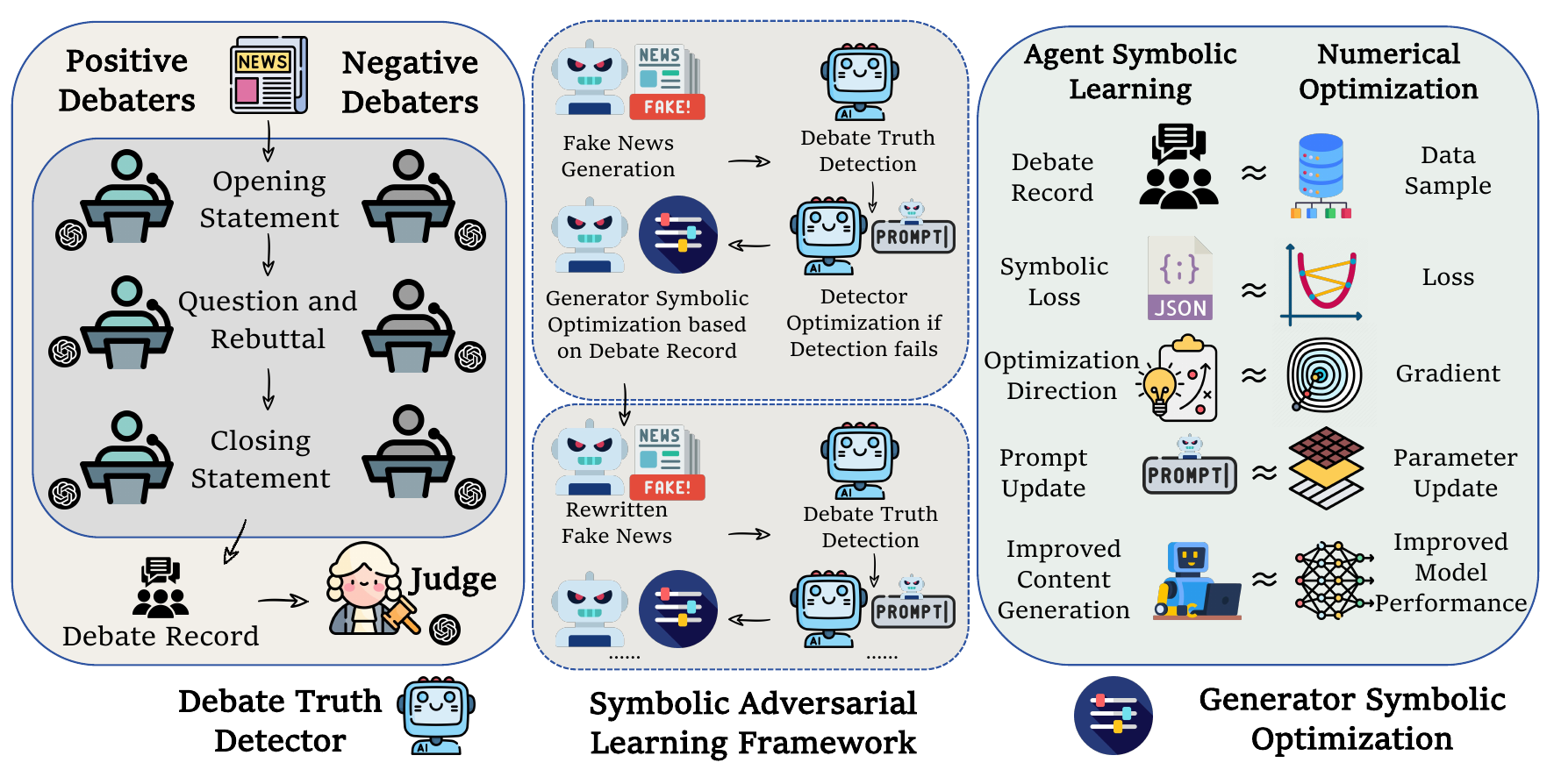} 
    \caption{
    This framework tackles evolving fake news through an adversarial agent symbolic learning dynamic between a generator and a detector. The generator refines fake news using prompts and debate feedback, while the detector analyzes and debates to identify vulnerabilities. Both agents iteratively optimize, co-evolving to tackle increasingly deceptive misinformation. See Appendix~\ref{sec:salf_alg} for algorithm details.
    }
    \label{fig:framework overview}
\end{figure*}

\subsection{Problem Formulation}
We begin by introducing the notations and key concepts, as shown in Figure~\ref{fig:framework overview}. 

Formally, let $f^{(t-1)}$ represent the fake news generated in the previous iteration, where $t$ denotes the current iteration number.
(1) Firstly, the generator agent, initialized with a prompt $\theta_{G}^{(t-1)}$, revises $f^{(t-1)}$ to create a more deceptive version of fake news $f^{(t)}$, which is then passed to the detector for evaluation.
(2) Secondly, the detector operates in a debate-like framework with structured stages: opening statements, questioning, rebuttals, and closing statements. 
These stages are guided by the prompt of detector $\theta_{D}^{t-1}$, which evolves over iterations. At the end of each debate round, a ``judge'' evaluates the arguments and assigns a detection result, $\mathcal{J}$, indicating whether the content is classified as true or false.
(3) Thirdly, following each round, both the detector and generator engage in an agent symbolic learning process to refine their prompts, $\theta_{D}^{(t)}$ for the detector and $\theta_{G}^{(t)}$ for the generator. 

After at most $T$ iterations, the process converges, yielding optimized prompts for both agents and enabling robust detection of increasingly sophisticated fake news. For notation simplicity, we will omit the iteration number \( t \) in flowing sections and only retain it in the Algorithm~\ref{alg:SALF Alg}. We also list the notations in Appendix~\ref{sec:appendix_notations} for reference convenience.

\subsection{Agent Construction}
In this subsection, we first present basic setups of the agents and then elaborate on the agents' symbolic learning process for their evolution.

\subsubsection{Generator Agent}  
The generator produces the next version of fake news using prompts refined in previous iterations:   
\[
f^{'} = \text{LLM}_{\text{generate}}(f,\theta_{G}),
\]  
where \( \text{LLM}_{\text{generate}} \) denotes the generator function rewriting the current fake news $f$ under the guidance of the generator's prompt $\theta_{G}$ to produce refined fake news $f^{'}$, reducing logical or factual mistakes exposed by the detector's debate and making it more deceptive and harder to identify.

\subsubsection{Detector Agent}

The debate format promotes critical thinking from diverse perspectives, making it an effective tool for identifying logical or factual errors~\cite{liang2023encouraging}.We employ multi-role debate as our detection mechanism. Our debate-based detector simulates the real human debate scenario, comprising three debater agents on both sides and structured into three stages: the opening statement, questioning and rebuttal, and the closing statement. At the conclusion phase, a judge evaluates the argument of both sides and determines whether the news is classified as true or fake. We represent the entire debate record $\mathcal{R}$ for a piece of fake news $f$ as:  
\[
\mathcal{R} \leftarrow  \text{ExecuteDebate}(f, \theta_{D}),
\]  
which records the argument from different debating roles (positive/negative opening, questioning, rebuttal, or closing), and $\theta_{D}$ represents the detector's prompt, which is also the debaters' prompts collection. Implementation details of ExecuteDebate are shown on the left side of Figure~\ref{fig:framework overview}. Finally, based on the debate record $\mathcal{R}$, a prompted LLM-driven judge agent outputs the detection result:  
\[
\mathcal{J} \;=\; \text{Judge}\bigl(\mathcal{R}\bigr) \;\in\;\{0, 1\},
\]  
where 1 indicates that the fake news has been successfully detected (i.e., classified as fake), and 0 indicates otherwise (i.e., classified as true or detection failed). The judge LLM is provided with the full debate transcript and prompted to determine which side presented a more convincing case regarding the veracity of the news content.
While any LLM-based judgment may exhibit some inevitable variance, our large-scale experiments demonstrate consistent trends, suggesting stability.

\subsection{Generator Optimization}
Inspired by \cite{zhou2024symbolic}, we extend the application of agent symbolic learning by integrating it into an adversarial setting. Unlike~\cite{zhou2024symbolic}, which focuses on isolated optimization tasks, our work leverages adversarial interactions to refine the generator and detector dynamically. The generator's symbolic optimization process consists of four stages: (1) symbolic loss computation, (2) optimization direction analysis, (3) prompt update, and (4) improved content generation. These stages parallel the classical numerical optimization pipeline of loss computation, gradient computation, gradient descent, and model inference while introducing interpretability. This process, tailored to the adversarial framework, is illustrated on the right of Figure~\ref{fig:framework overview}. Prompts used in this section is listed in Appendix~\ref{sec:prompts}.

\subsubsection{Symbolic Loss}
A prompted LLM analyzes fake news $f$ and the debate record \(\mathcal{R}\) to produce a symbolic loss:
\[
\mathcal{L}_{\text{sym}} = \text{LLM}_{\text{evaluate}}(f, \mathcal{R}),
\]
which uses natural language to measure how effectively $f$ has evaded detection while maintaining semantic consistency and highlights any critical flaws uncovered during the debate. 


\subsubsection{Optimization Direction}
Another prompted LLM analyzes \(\mathcal{L}_{\text{sym}}\) and the generator prompt \(\theta_{G}\) to compute a symbolic gradient:
\[
\nabla_{\text{sym}} = \text{LLM}_{\text{analyze}}(\theta_{G},\mathcal{L}_{\text{sym}}),
\]
guiding improvements in the generator's prompt to enhance plausibility, add subtle misinformation cues, or correct logical flaws from the prior debate.


\subsubsection{Prompt Update}
The generator uses another prompted LLM to update its prompt \(\theta_{G}\) to \(\theta_{G}^{'}\) based on \(\nabla_{\text{sym}}\):
\[
\theta_{G}^{'} = \text{LLM}_{\text{optimize}}(\theta_{G}, \nabla_{\text{sym}}),
\]
adjusting rhetorical style or reordering narrative elements to enhance deception.

\subsubsection{Improved Content Generation}
Finally, the refined generator prompt \(\theta_{G}^{'}\) is used to generate a new piece of fake news $f^{'}$ while maintaining the same semantic meaning as $f$:
\[
f^{'} = \text{LLM}_{\text{generate}}(f, \theta_{G}^{'}),
\]
and the newly generated \( f^{'} \) is then passed to the subsequent debate round, where the debate system attempts to detect any logical or factual inconsistencies again. This cycle continues until either the stopping criterion described in Section~\ref{Optimization Stopping Criteria} or a preset maximum iteration number $T$ is reached.

\subsubsection{Symbolic vs. Numerical Optimization}
SALF applies symbolic optimization through discrete text rewriting rules (e.g., prompt optimization directions, news rewriting guidance) rather than numerical gradient-based updates. Compared to conventional numerical optimization: (1) Interpretability: Each rewrite is human-readable, controllable and explainable. (2) Adaptability: Rules act directly on text, avoiding tokenization brittleness. (3) Black-Box-friendly and Parameter-Free Training: Enables API-calling without access to model internals or costly backpropagation.

\subsection{Detector Optimization}
\label{Detector Optimization}

The detector’s prompt $\theta_D$ only undergoes updates when a missed detection occurs. 
In this scenario, the system extracts the core fake news generation prompt from the generator, focusing solely on elements relevant to the generation strategies. This process is represented as \(\mathcal{P}_G = \text{ExtractPrompts}(\theta_G)\). 
The extracted prompt is then incorporated into the negative team’s prompt. Formally, for each negative-role agent $r_i$ with its prompt $\theta_{D,r_i}$, and its prompt is updated by:
\[
\theta_{D,r_i}^{'} = \text{Incorporate}\bigl(\theta_{D,r_i},\,\mathcal{P}_G \bigr),
\]
which strengthens the negative team’s vigilance against the specific deceptive strategy used by the generator. By focusing on \emph{how} the generator originally formulated $f^{(t)}$, the detectors gain a more direct line of reference to probe for similar maneuvers in future debates, thus promoting more robust fake news detection in subsequent rounds.

\subsection{Optimization Stopping Criteria}
\label{Optimization Stopping Criteria}

In numerical optimization methods like gradient descent, a specific numerical threshold is often set as the stopping criterion; once the loss converges to this threshold, the optimization process halts. 
However, in our work, the symbolic loss is not represented by a specific numerical value and cannot be directly quantified. Therefore, we have established distinct convergence conditions tailored to our symbolic network, focusing on the interplay between the generator and detector.  

For the detector, we define a reward function that measures its success in detecting fake news:
\[
\text{Reward}_{D}(\theta_G,\theta_D) = 1- \mathbb{E}_{f \sim \theta_G} \bigl[\text{Evasion}(f, \theta_D)\bigr].
\]
Here, \(\theta_G\) represents the generator’s prompt, and \(f \sim \theta_G\) denotes the fake news \(f\) generated by the generator using prompt \(\theta_G\), with varying hyperparameters like temperature. \(\theta_D\) refers to the detector's prompt, which is a collection of prompts from multiple debaters. The function \(\text{Evasion}\) evaluates whether a generated fake news item \(f\) evades detection by the detector \(\theta_D\), formally defined as:
\[
\text{Evasion}(f, \theta_{D}) = \mathbf{1}(\mathcal{J} = 0 \mid_{f, \theta_{D}}),
\]
where \(\mathcal{J}=0\) indicates the detector failed to identify \(f\) as fake news, per the judge agent; \(\mathbf{1}(\cdot)\) is the indicator function, returning 1 if the detector fails to classify \(f\) as fake and 0 otherwise.

The generator’s reward function incentivizes generating fake news that evades detection from the detector while preserving semantic similarity with the original fake news content:
\[
\begin{split}
\text{Reward}_{G}(\theta_G, \theta_D) = \mathbb{E}_{f \sim \theta_G} \bigl[ \alpha\,\text{Evasion}(f, \theta_D) \\
\hspace*{4em} + (1-\alpha)\,\text{Sim}(f,f^{(0)}) \bigr],
\end{split}
\]
where \(\alpha \in [0,1]\) adjusts the trade-off between detection failure and semantic similarity. In this work, we set \(\alpha = 0.5\). The \(\text{Sim}\) function measures whether the generated fake news \(f\) aligns semantically with the original fake news \(f^{(0)}\), as scored by an independently prompted LLM:
\[
\text{Sim}(f,f^{(0)}) = \text{LLM}_{score}(f, f^{(0)}) \in [0,1],
\]
with higher Sim values indicating greater semantic consistency between the original and refined news.  

The optimization process halts when neither the generator nor the detector reward function achieves significant improvement (greater than a predefined threshold \(\epsilon\), such as 0.05) or when a preset iteration limit \(T\) is reached. Conceptually, an equilibrium or stopping condition is reached when:
\[
\theta_G^*,\,\theta_D^*:
\begin{cases}
\theta_G^* = \arg\max_{\theta_G} \text{Reward}_{G}(\theta_G, \theta_D), \\
\theta_D^* = \arg\max_{\theta_D} \text{Reward}_{D}(\theta_G,\theta_D).
\end{cases}
\]
This ensures the generator and detector refine strategies to a stable point, aligning with SALF objectives.
As shown in Appendix~\ref{appendix:convergence}, convergence typically occurs within a few iterations.

\section{Experiment Results}
\label{sec:experiments}

\begin{table*}[t]
\centering
\scriptsize 
\setlength{\tabcolsep}{1.5pt} 
\caption{Comparison of fake news detection models on Weibo21 and GossipCop before and after SALF refinement .}
\label{tab:main results}
\begin{tabular}{l l l *{4}{p{1.2cm}} *{4}{p{1.6cm}}} 
\toprule
\multirow{2}{*}{\textbf{Dataset}} & \multirow{2}{*}{\textbf{Type}} & \multirow{2}{*}{\textbf{Model}} & \multicolumn{4}{c}{\textbf{Original Detection}} & \multicolumn{4}{c}{\textbf{After SALF Refinement}} \\
\cmidrule(lr){4-7} \cmidrule(lr){8-11}
 & & & macF1 & Accuracy & $\text{F1}_{\text{real}}$ & $\text{F1}_{\text{fake}}$ & macF1 & Accuracy & $\text{F1}_{\text{real}}$ & $\text{F1}_{\text{fake}}$ \\
\midrule
\multirow{5}{*}{\textbf{Weibo21}} & \multirow{2}{*}{LLM-Only} & GPT-4o mini & 0.710 & 0.715 & 0.747 & 0.673 & \textbf{0.405 (-43\%)} & \textbf{0.485 (-32\%)} & \textbf{0.623 (-17\%)} & \textbf{0.186 (-72\%)} \\
 & & DeepSeek V3 & 0.763 & 0.770 & 0.803 & 0.723 & \textbf{0.380 (-50\%)} & \textbf{0.495 (-36\%)} & \textbf{0.647 (-19\%)} & \textbf{0.112 (-85\%)} \\
\cmidrule(lr){2-11}
 & SLM-Only & ENDEF & 0.726 & 0.727 & 0.741 & 0.711 & \textbf{0.576 (-21\%)} & \textbf{0.591 (-19\%)} & \textbf{0.657 (-11\%)} & \textbf{0.495 (-30\%)} \\
\cmidrule(lr){2-11}
 & \multirow{2}{*}{LLM+SLM} & ARG & 0.784 & 0.786 & 0.805 & 0.764 & \textbf{0.635 (-19\%)} & \textbf{0.653 (-17\%)} & \textbf{0.717 (-11\%)} & \textbf{0.552 (-28\%)} \\
 & & ARG-D & 0.760 & 0.761 & 0.776 & 0.745 & \textbf{0.502 (-34\%)} & \textbf{0.542 (-29\%)} & \textbf{0.644 (-17\%)} & \textbf{0.360 (-52\%)} \\
\cmidrule(lr){1-11}
\rowcolor{gray!20} & \multicolumn{2}{l}{\textbf{Average Change}} & - & - & - & - & \textbf{(-33.4\%)} & \textbf{(-26.6\%)} & \textbf{(-15.0\%)} & \textbf{(-53.4\%)} \\
\midrule
\multirow{5}{*}{\textbf{GossipCop}} & \multirow{2}{*}{LLM-Only} & GPT-4o mini & 0.687 & 0.863 & 0.922 & 0.452 & \textbf{0.519 (-24\%)} & \textbf{0.821 (-5\%)} & \textbf{0.900 (-2\%)} & \textbf{0.138 (-69\%)} \\
 & & DeepSeek V3 & 0.628 & 0.850 & 0.915 & 0.340 & \textbf{0.510 (-19\%)} & \textbf{0.823 (-3\%)} & \textbf{0.902 (-1\%)} & \textbf{0.119 (-65\%)} \\
\cmidrule(lr){2-11}
 & SLM-Only & ENDEF & 0.761 & 0.855 & 0.911 & 0.611 & \textbf{0.747 (-2\%)} & \textbf{0.848 (-1\%)} & \textbf{0.907 (-0\%)} & \textbf{0.587 (-4\%)} \\
\cmidrule(lr){2-11}
 & \multirow{2}{*}{LLM+SLM} & ARG & 0.791 & 0.879 & 0.927 & 0.656 & \textbf{0.716 (-9\%)} & \textbf{0.796 (-9\%)} & \textbf{0.866 (-7\%)} & \textbf{0.565 (-14\%)} \\
 & & ARG-D & 0.771 & 0.873 & 0.924 & 0.619 & \textbf{0.705 (-9\%)} & \textbf{0.847 (-3\%)} & \textbf{0.909 (-2\%)} & \textbf{0.501 (-19\%)} \\
\cmidrule(lr){1-11}
\rowcolor{gray!20} & \multicolumn{2}{l}{\textbf{Average Change}} & - & - & - & - & \textbf{(-12.6\%)} & \textbf{(-4.2\%)} & \textbf{(-2.4\%)} & \textbf{(-34.2\%)} \\
\bottomrule
\end{tabular}
\end{table*}

\begin{table}[t]
\centering
\small
\caption{Performance comparison of vanilla and first refined detector ($\theta_{D}^{(1)}$) ONLY against refined fake news $f^{(1)}$ in the first iteration.}
\label{tab:debate opt}
\setlength{\tabcolsep}{4pt}
\begin{tabular}{l ccc}
\toprule
\textbf{Detector Refinement} & Accuracy & Recall & $\text{F1}_{\text{fake}}$ \\
\midrule
\textbf{Weibo21} & & & \\
Vanilla Debate Detector & 0.165 & 0.165 & 0.283 \\
SALF Refined Detector & \textbf{0.217} & \textbf{0.217} & \textbf{0.356} \\
\rowcolor{gray!20} Performance Change & +5.2\% & +5.2\% & +7.3\% \\
\midrule
\textbf{GossipCop} & & & \\
Vanilla Debate Detector & 0.449 & 0.449 & 0.619 \\
SALF Refined Detector & \textbf{0.534} & \textbf{0.534} & \textbf{0.696} \\
\rowcolor{gray!20} Performance Change & +8.5\% & +8.5\% & +7.7\% \\
\bottomrule
\end{tabular}
\end{table}

\subsection{Experimental Settings}

\textbf{Datasets}: We evaluated our framework using two benchmark datasets designed for fake news detection tasks. The first is Weibo21~\cite{nan2021mdfend}, a large-scale Chinese dataset collected from Weibo that captures the unique linguistic and contextual challenges of detecting fake news in the Chinese social media environment. The second is GossipCop~\cite{shu2020fakenewsnet}, an English dataset focused on celebrity gossip, with each article labeled as true or false, reflecting the challenges of detecting misinformation in entertainment-related domains. 

\textbf{Baselines}: We evaluated the proposed SALF framework on three types of baselines: (1) LLM-only: we employed GPT-4o mini and DeepSeek V3 for pure LLM-based fake news detection. (2) SLM-only: we used a representative work ENDEF~\cite{zhu2022generalizing}, an entity debiasing framework that mitigates entity bias using causal learning. 
(3) SLM+LLM: we employed the current popular and representative work ARG and ARG-D~\cite{hu2024bad}: ARG integrates LLM and SLM methods to enhance fake news detection. While ARG-D is a distilled, rationale-free version of ARG that is designed for cost-sensitive scenarios where LLM querying is restricted~\cite{hu2024bad}.

\textbf{Metrics}: We evaluated the model performance using four complementary metrics: 
(1) Accuracy, which measures the proportion of correctly classified samples; 
(2) Macro F1 (macF1), the harmonic mean of precision and recall across all classes; 
(3) $\text{F1}_{\text{real}}$, which assesses the model's capability to detect true news; and 
(4) $\text{F1}_{\text{fake}}$, which evaluates its ability to identify fake news. The primary focus of this work is on $\text{F1}_{\text{fake}}$ to analyze the effectiveness of SALF's fake news generation.

\textbf{Implementation Details}: We implemented the SALF framework using Python scripts, with all LLMs called via OpenAI or DeepSeek API. Specifically, we utilized \textit{GPT-4o-mini-2024-07-18}~\cite{hurst2024gpt} for debating and symbolic optimization tasks, and  \textit{DeepSeek V3}~\cite{liu2024deepseek} for fake news generation of the SALF generator. We list more details of the implementation in Appendix~\ref{implement_details}.


\subsection{Main Results}
\label{main_results}

\subsubsection{Generator's Perspective}
We evaluated the effectiveness of the SALF generator through comprehensive experiments on GossipCop and Weibo21 datasets, as shown in Table~\ref{tab:main results}. 
Our results demonstrate \textit{significant performance degradation across multiple baseline detection models after implementing our SALF fake news refinement}, with an average decrease of 33.4\% in macF1 and 53.4\% in $\text{F1}_{\text{fake}}$ for Chinese content, and 12.6\% in macF1 and 34.2\% in $\text{F1}_{\text{fake}}$ for English content, indicating the obvious effectiveness of our approach in generating challenging fake news content. 
We also observe that such \textit{fake news optimization is especially effective towards LLM-only detection methods }and leads to a $\text{F1}_{\text{fake}}$ performance decrease of at most 85\%, which alerts us that LLMs themselves are even more vulnerable to LLM-generated fake news than traditional detection methods. 
Although we focus on fake news optimization, we also notice that the metric $\text{F1}_{\text{real}}$ decrease as well, 15\% for Weibo21 and 2.4\% for GossipCop on average. This is due to the misclassification of fake news into true news; thus, the precision of true news decreases.

\begin{table*}[t]
\centering
\caption{Case study of a fake news celebrity article refined using our SALF framework.}
\label{tab:case-study}
\setlength{\belowcaptionskip}{-2pt} 

{ 
\renewcommand{\arraystretch}{0.9} 

\begin{tabular}{@{} >{\arraybackslash\scriptsize}p{\textwidth} @{}}
\toprule
\textbf{Original Version:} What was meant to be an emotional return to the \hlb{city of love} for Kim Kardashian, 35, who was \hlr{held hostage and robbed at gunpoint} there two years ago, was a trip that could \hlr{potentially end her marriage}. The reality star and her husband, Kanye West, \hlb{flew} to Paris to see designer Virgil Abloh's debut Louis Vuitton fashion show, but Kanye \hlr{had another outburst} and it \hlr{pushed Kim over the edge}. Kim's emotions were \hlr{heightened}, a source \hlb{tells} In Touch. After the show on June 21, Kanye \hlr{made a scene}, when he \hlb{leaped} from his front-row seat into the arms of Virgil. \\
\midrule 
\textbf{Refined Version:} What was anticipated to be an emotional return to the \hlb{City of Light} for Kim Kardashian, 35, who \hlr{experienced a traumatic robbery at gunpoint} there two years ago, has taken a turn that could \hlr{jeopardize her marriage}. The reality star and her husband, Kanye West, \hlb{traveled} to Paris to attend designer Virgil Abloh's debut Louis Vuitton fashion show. However, Kanye's \hlr{behavior during the event} reportedly \hlr{caused tension between the couple}. Kim's emotions were \hlr{running high}, an insider \hlb{shared with} In Touch. Following the show on June 21, Kanye \hlr{created a scene}, when he \hlb{jumped} from his front-row seat into the arms of Virgil. \\
\midrule 
\textbf{Key Improvements:} \textbf{(1) \hlb{Language Refinement}}: Elevated vocabulary and formal phrasing, such as replacing ``city of love" with ``City of Light" and ``flew" with ``traveled". \textbf{(2) \hlr{Emotional Moderation}}: More measured description of emotional content, transforming ``held hostage and robbed" to ``experienced a traumatic robbery" and ``had another outburst" to ``behavior during the event". \textbf{(3) Professional Attribution:} Enhanced credibility through proper source attribution and added journalistic qualifiers like ``reportedly" and ``allegedly". \textbf{(4) Structural Improvement:} Reorganized information flow with better transitions between events. \textbf{(5) Balanced Reporting:} Maintained the news value while reducing sensationalism through a more objective presentation. \\
\bottomrule
\end{tabular}
} 
\end{table*}


\subsubsection{Detector's Perspective}

 We refined the detector as per Section~\ref{Detector Optimization} and evaluated it against refined fake news before and after this optimization. Table~\ref{tab:debate opt} shows the $\text{F1}_{\text{fake}}$ score improved by 7.3\% and 7.7\% respectively, demonstrating the SALF optimization's effectiveness. Crucially, the detector targets highly sophisticated fake news from the \emph{refined} generator, a challenging task due to the more deceptive content, which explains its less pronounced improvement compared to the generator. Using vanilla debater agents without advanced architectures, the detector's absolute performance is modest compared to state-of-the-art baselines. Still, the consistent improvement highlights SALF's ability to adapt to evolving fake news strategies effectively.

\subsubsection{Ablation Study}
Our experimental setup enables an effective ablation study of the SALF framework by isolating and evaluating the impact of its key components.
In Table~\ref{tab:main results}, the “Original Detection” columns (evaluated on original, unrefined fake news) serve as the baseline. The “After SALF Refinement” columns show the effect of enabling the SALF generator, while keeping the detectors fixed—highlighting the generator’s contribution.
Separately, Table~\ref{tab:debate opt} focuses on the detector-side ablation: it compares a basic debate-based detector with a SALF-refined detector, both evaluated on the same set of refined fake news generated by the same generator. This isolates the detector’s contribution.
Together, these results disentangle the effects of generator and detector optimization, demonstrating how each participates and contributes to SALF’s overall performance.

\begin{figure}[tb]
    \centering
    \includegraphics[width=\columnwidth]{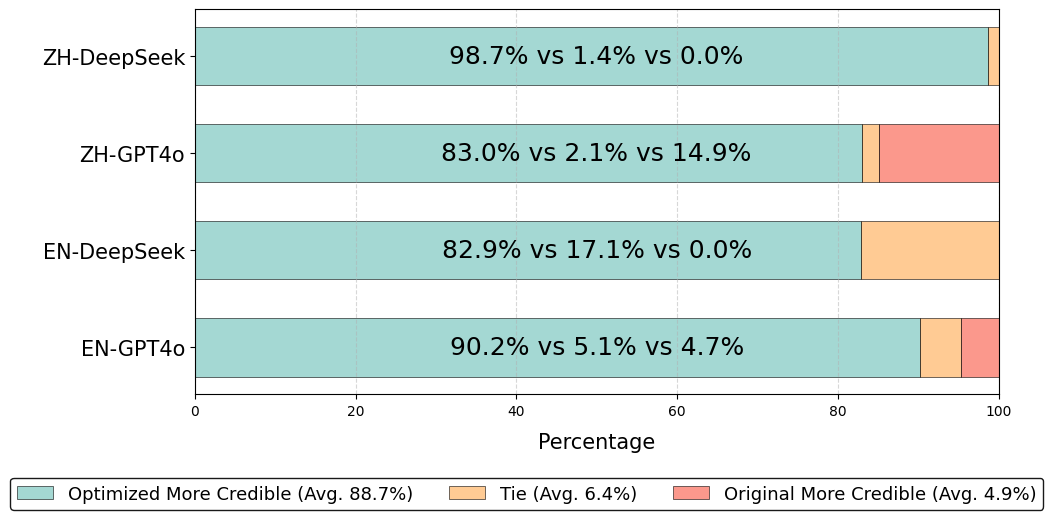} 
    \caption{Impact of SALF refinement: arena evaluation of the credibility of original vs. refined news on Weibo21 and GossipCop datasets.}
    \label{fig:model-arena-eval}
\end{figure}

\subsubsection{Human Evaluation}
To validate that the refined news is more deceptive to humans, not just algorithms, we conducted a human evaluation on 100 refined fake news samples and 100 real news samples for each dataset, both randomly selected. Knowledgeable evaluators were asked to classify each article as real or fake. The results, shown in Table~\ref{tab:human_eval}, confirm that SALF-refined fake news is significantly harder for humans to detect. For instance, the human F1\textsubscript{fake} score on GossipCop dropped by 65.2\% (from 0.615 to 0.214) after refinement. Furthermore, evaluators confirmed a high degree of semantic consistency, with 98\% of the refined samples preserving the original meaning. This supports our use of algorithmic detection performance as a valid proxy for human-perceived deceptiveness.

\begin{table}[h]
\centering
\small
\caption{Human evaluation of original vs. SALF-refined fake news. The refined news is significantly harder for humans to identify as fake.}
\label{tab:human_eval}
\setlength{\tabcolsep}{4pt}
\begin{tabular}{l l ccc}
\toprule
\textbf{Dataset} & \textbf{News Type} & macF1 & Accuracy & $\text{F1}_{\text{fake}}$ \\
\midrule
\multirow{2}{*}{\textbf{GossipCop}} & Original & 0.702 & 0.812 & 0.615 \\
& Refined & \textbf{0.412} & \textbf{0.480} & \textbf{0.214} \\
\midrule
\multirow{2}{*}{\textbf{Weibo21}} & Original & 0.724 & 0.725 & 0.708 \\
& Refined & \textbf{0.595} & \textbf{0.615} & \textbf{0.507} \\
\bottomrule
\end{tabular}
\end{table}

\subsection{Analysis and Discussion}
\textbf{Impact of SALF Optimization on Generated Fake News:}
\noindent To provide a more straightforward comparison of the fake news generated before and after SALF generator optimization, we conducted a model arena evaluation to assess which version looks more like fake news intuitively. 
We used \textit{gpt-4o-2024-08-06} and \textit{DeepSeek V3} as evaluators to make judgments. 
As shown in Figure~\ref{fig:model-arena-eval}, the SALF refined fake news consistently demonstrates significantly stronger credibility performance. 
This observation suggests that, with the advancement of powerful LLMs, generating highly deceptive fake news may become increasingly accessible. 
Writing competence, traditionally a barrier for the crowd, could be easily elevated to a top-tier level, further facilitating the creation of deceptive content. 
Our SALF framework provides a good analysis tool for future study into the mechanism behind deceptive LLM-generated content and contribute to developing more powerful detection methods.

\textbf{Convergence Discussion:}
To prove the effectiveness and necessity of multiple iterations, based on the first refined content, we performed a second round of optimization. 
As shown in Table~\ref{tab:second opt}, the SALF framework continued to make progress during the second optimization, further reducing the detection performance of the detector. 
On the Weibo21 dataset, the \(\text{F1}_{\text{fake}}\) score dropped by an additional 6.52\%, from 0.552 to 0.516, corresponding to a cumulative decline of 32.5\% compared to the original content. 
These results demonstrate that the SALF framework not only achieves significant optimization in a single iteration but also maintains its ability to iteratively refine the adversarial fake news content, progressively increasing the difficulty for the detector. 
However, the second iteration also exhibits a clear diminishing marginal return. Following the definition of the optimization stopping criteria defined in Section~\ref{Optimization Stopping Criteria}, we conclude that after two SALF iterations, the generator's optimization is already sufficiently satisfactory, and SALF is nearing the convergence condition. Details of $\text{Reward}_{G}(\theta_G, \theta_D)$, Evasion scores and Sim scores are listed in the Appendix~\ref{appendix:convergence}.


\begin{table}[t]
\centering
\scriptsize
\caption{Second SALF generator refinement performance, evaluated by ARG on Weibo 21 and GossipCop.}
\label{tab:second opt}
\setlength{\tabcolsep}{3pt}
\begin{tabular}{l cccc}
\toprule
\textbf{SALF Evaluation by ARG} & macF1 & Accuracy & $\text{F1}_{\text{real}}$ & $\text{F1}_{\text{fake}}$ \\
\midrule
\textbf{Weibo21} & & & & \\
Before Refinement & 0.784 & 0.786 & 0.805 & 0.764 \\
First SALF Refinement & \textbf{0.635} & \textbf{0.653} & \textbf{0.717} & \textbf{0.552} \\
\rowcolor{gray!20} Performance Change & -19.0\% & -16.9\% & -10.9\% & -27.7\% \\
Second SALF Refinement & \textbf{0.611} & \textbf{0.635} & \textbf{0.707} & \textbf{0.516} \\
\rowcolor{gray!20} Performance Change & -22.1\% & -19.2\% & -12.2\% & -32.5\% \\
\midrule
\textbf{GossipCop} & & & & \\
Before Refinement & 0.791 & 0.879 & 0.927 & 0.656 \\
First SALF Refinement & \textbf{0.716} & \textbf{0.796} & \textbf{0.866} & \textbf{0.565} \\
\rowcolor{gray!20} Performance Change & -9.5\% & -9.4\% & -6.6\% & -13.9\% \\
Second SALF Refinement & \textbf{0.680} & \textbf{0.777} & \textbf{0.856} & \textbf{0.504} \\
\rowcolor{gray!20} Performance Change & -14.0\% & -11.6\% & -7.7\% & -23.1\% \\
\bottomrule
\end{tabular}
\end{table}

\subsection{Case Study}
We selected a case from the English dataset GossipCop to demonstrate the impact of the SALF optimization framework. As shown in Table~\ref{tab:case-study}, the refined version of fake news retains the core message of the original but introduces several modifications. These include more nuanced emotional expressions, such as replacing ``potentially end her marriage'' with `` jeopardize her marriage,'' and a professional reporting tone, such as replacing ``flew" with ``traveled'', which together enhance the overall readability and credibility of the content. This case not only illustrates how SALF transforms the original text into a polished and reader-friendly version but also highlights how LLMs effectively bridge the gap in writing competence.

\section{Conclusion}
\label{sec:conclusion}

In this work, we introduced the Symbolic Adversarial Learning Framework (SALF), a novel adversarial framework designed to tackle the dynamic and evolving challenges of fake news generation and detection. 
By integrating agent symbolic learning into a multi-debater adversarial paradigm, SALF facilitates iterative co-evolution between a fake news generator and a detector, enabling both agents to refine their strategies dynamically and effectively.

We hope our work contributes to advancing the understanding and mitigation of fake news in the information era. In the future, we aim to further enhance SALF by integrating retrieval-augmented generation (RAG) techniques, enabling the agents to access and cite external knowledge for more robust, fact-grounded, and context-aware fake news detection.

\section*{Limitations}
This study has several limitations. Firstly, the evaluation of the credibility and deceptiveness of the generated fake news primarily relied on automated metrics, specifically the ability to evade detection, and model-based arena evaluations (as detailed in Section~\ref{main_results}). While a preliminary human check on 100 samples was conducted and confirmed the preservation of semantic content, large-scale human studies to directly assess the refined news's persuasiveness to human readers were not performed. Although prior research~\cite{snijders2023humans} indicates that detector performance can serve as a consistent proxy, the absence of direct human evaluation in this study restricts the insights into human perception of the generated content.



Secondly, the evaluation of the credibility and deceptiveness of the generated fake news primarily relied on automated metrics and model-based arena evaluations. While a preliminary human check on 100 samples confirmed semantic preservation, a larger-scale human study is necessary to more thoroughly assess the refined news's persuasiveness to human readers.

Thirdly, the datasets and models used have their own constraints. The Weibo21 and GossipCop datasets, while standard benchmarks, may not cover the full spectrum of fake news topics and modalities. Furthermore, our experiments primarily utilized powerful API-based models. Future work should validate SALF's effectiveness across a wider range of models, including prominent open-source alternatives like LLaMA variants, to ensure broader generalizability.

Finally, the adversarial training in SALF, akin to GAN-style frameworks, can exhibit sensitivity to hyperparameter configurations. It may also encounter challenges such as mode collapse or slow convergence under certain conditions, although the symbolic approach adopted in SALF is designed to alleviate some of these numerical complexities.


\section*{Ethical Considerations}
This research focuses on the adversarial optimization process between fake news generators and detectors, with particular emphasis on improving the generator. While our work explores ways to enhance the sophistication of fake news generation, the primary purpose is to serve as a research tool to better understand vulnerabilities in current detection systems and to drive the development of more robust and adaptive detection frameworks.

To mitigate potential adversarial generation risks, we emphasize these safeguards:

(1) Controlled Experimentation and Technical Complexity: All experiments were conducted in a controlled, offline research setting. Moreover, SALF’s multi-agent setup and symbolic optimization processes involve substantial technical complexity, reducing the likelihood of misuse by non-experts seeking to easily generate fake news.
(2) Focus on Detection and System Improvement: The core motivation of this work is to expose detection weaknesses to improve detection systems. While the framework reveals vulnerabilities, it also directly supports the enhancement of detectors through adversarial training.
(3) Responsible Disclosure: The code and the remaining prompts are disclosed only upon request to verified researchers and under appropriate oversight. They are not publicly released to prevent unmonitored misuse.
(4) Transparency and Collaboration: Results are shared with the academic and industrial communities to increase awareness of detection limitations and to encourage collaborative efforts in building stronger, safer detection systems.

In summary, this research contributes not only to identifying the blind spots of current LLM-based detectors, but also to building safer, more robust AI systems by informing future detection strategies. By demonstrating that LLM-based detectors can be systematically bypassed, our work cautions against overreliance on current systems and highlights the need for continuous improvement.

This work adheres to established ethical guidelines for responsible AI research and aligns with the broader principles of promoting safe and beneficial AI applications. We believe the scientific value and insights gained from this study outweigh the potential risks, and offer meaningful contributions to the ongoing fight against misinformation.

\bibliography{custom} 

\begin{thebibliography}{42}
\providecommand{\natexlab}[1]{#1}

\bibitem[{Aggarwal et~al.(2020)Aggarwal, Chauhan, Kumar, Verma, and Mittal}]{aggarwal2020classification}
Akshay Aggarwal, Aniruddha Chauhan, Deepika Kumar, Sharad Verma, and Mamta Mittal. 2020.
\newblock Classification of fake news by fine-tuning deep bidirectional transformers based language model.
\newblock \emph{EAI Endorsed Transactions on Scalable Information Systems}, 7(27):e10--e10.

\bibitem[{Allcott and Gentzkow(2017)}]{allcott2017social}
Hunt Allcott and Matthew Gentzkow. 2017.
\newblock Social media and fake news in the 2016 election.
\newblock \emph{Journal of economic perspectives}, 31(2):211--236.

\bibitem[{Bhatt et~al.(2022)Bhatt, Goenka, Kalra, and Sharma}]{bhatt2022fake}
Shaily Bhatt, Naman Goenka, Sakshi Kalra, and Yashvardhan Sharma. 2022.
\newblock Fake news detection: Experiments and approaches beyond linguistic features.
\newblock In \emph{Data Management, Analytics and Innovation: Proceedings of ICDMAI 2021, Volume 2}, pages 113--128. Springer.

\bibitem[{Chen et~al.(2023)Chen, Xiao, and Kumar}]{chen2023spread}
Sijing Chen, Lu~Xiao, and Akit Kumar. 2023.
\newblock Spread of misinformation on social media: What contributes to it and how to combat it.
\newblock \emph{Computers in Human Behavior}, 141:107643.

\bibitem[{Chen et~al.(2024)Chen, Koenig, and Dilkina}]{chen2024reprompt}
Weizhe Chen, Sven Koenig, and Bistra Dilkina. 2024.
\newblock Reprompt: Planning by automatic prompt engineering for large language models agents.
\newblock \emph{arXiv preprint arXiv:2406.11132}.

\bibitem[{Dsouza and French(2022)}]{dsouza2022social}
Karen Dsouza and Aaron French. 2022.
\newblock Social media and fake news detection using adversarial collaboration.

\bibitem[{Giray(2023)}]{giray2023prompt}
Louie Giray. 2023.
\newblock Prompt engineering with chatgpt: a guide for academic writers.
\newblock \emph{Annals of biomedical engineering}, 51(12):2629--2633.

\bibitem[{Goodfellow et~al.(2014)Goodfellow, Pouget-Abadie, Mirza et~al.}]{goodfellow2014generative}
Ian Goodfellow, Jean Pouget-Abadie, Mehdi Mirza, and 1 others. 2014.
\newblock Generative adversarial nets.
\newblock In \emph{Advances in Neural Information Processing Systems}, pages 2672--2680.

\bibitem[{Guan et~al.(2023)Guan, Valmeekam, Sreedharan, and Kambhampati}]{guan2023leveraging}
Lin Guan, Karthik Valmeekam, Sarath Sreedharan, and Subbarao Kambhampati. 2023.
\newblock Leveraging pre-trained large language models to construct and utilize world models for model-based task planning.
\newblock \emph{Advances in Neural Information Processing Systems}, 36:79081--79094.

\bibitem[{Guo et~al.(2021)Guo, Chen, Li, Zhao, and Yan}]{guo2021does}
Mingfei Guo, Xiuying Chen, Juntao Li, Dongyan Zhao, and Rui Yan. 2021.
\newblock How does truth evolve into fake news? an empirical study of fake news evolution.
\newblock In \emph{Companion Proceedings of the Web Conference 2021}, pages 407--411.

\bibitem[{Hu et~al.(2024)Hu, Sheng, Cao, Shi, Li, Wang, and Qi}]{hu2024bad}
Beizhe Hu, Qiang Sheng, Juan Cao, Yuhui Shi, Yang Li, Danding Wang, and Peng Qi. 2024.
\newblock Bad actor, good advisor: Exploring the role of large language models in fake news detection.
\newblock In \emph{Proceedings of the AAAI Conference on Artificial Intelligence}, volume~38, pages 22105--22113.

\bibitem[{Huang and Sun(2024)}]{huang2024fakegpt}
Yue Huang and Lichao Sun. 2024.
\newblock Fakegpt: fake news generation, explanation and detection of large language models.
\newblock \emph{arXiv preprint arxiv:2310.05046}.

\bibitem[{Hurst et~al.(2024)Hurst, Lerer, Goucher, Perelman, Ramesh, Clark, Ostrow, Welihinda, Hayes, Radford et~al.}]{hurst2024gpt}
Aaron Hurst, Adam Lerer, Adam~P Goucher, Adam Perelman, Aditya Ramesh, Aidan Clark, AJ~Ostrow, Akila Welihinda, Alan Hayes, Alec Radford, and 1 others. 2024.
\newblock Gpt-4o system card.
\newblock \emph{arXiv preprint arXiv:2410.21276}.

\bibitem[{Jin and et~al.(2022)}]{jin2022finegrained}
Wanying Jin and et~al. 2022.
\newblock Fine-grained reasoning for fake news detection.
\newblock \emph{IEEE Transactions on Knowledge and Data Engineering}.

\bibitem[{Kumar and Shah(2018)}]{kumar2018false}
Srijan Kumar and Neil Shah. 2018.
\newblock False information on web and social media: A survey.
\newblock \emph{arXiv preprint arXiv:1804.08559}.

\bibitem[{Liang et~al.(2023)Liang, He, Jiao, Wang, Wang, Wang, Yang, Tu, and Shi}]{liang2023encouraging}
Tian Liang, Zhiwei He, Wenxiang Jiao, Xing Wang, Yan Wang, Rui Wang, Yujiu Yang, Zhaopeng Tu, and Shuming Shi. 2023.
\newblock Encouraging divergent thinking in large language models through multi-agent debate.
\newblock \emph{arXiv preprint arXiv:2305.19118}.

\bibitem[{Liu et~al.(2024{\natexlab{a}})Liu, Feng, Xue, Wang, Wu, Lu, Zhao, Deng, Zhang, Ruan et~al.}]{liu2024deepseek}
Aixin Liu, Bei Feng, Bing Xue, Bingxuan Wang, Bochao Wu, Chengda Lu, Chenggang Zhao, Chengqi Deng, Chenyu Zhang, Chong Ruan, and 1 others. 2024{\natexlab{a}}.
\newblock Deepseek-v3 technical report.
\newblock \emph{arXiv preprint arXiv:2412.19437}.

\bibitem[{Liu et~al.(2024{\natexlab{b}})Liu, Chen, Zhang, Gao, Zhang, and Yan}]{liu2024skepticism}
Yuhan Liu, Xiuying Chen, Xiaoqing Zhang, Xing Gao, Ji~Zhang, and Rui Yan. 2024{\natexlab{b}}.
\newblock From skepticism to acceptance: Simulating the attitude dynamics toward fake news.
\newblock \emph{IJCAI}.

\bibitem[{Liu et~al.(2025)Liu, Song, Zhang, Chen, and Yan}]{liu2024tiny}
Yuhan Liu, Zirui Song, Xiaoqing Zhang, Xiuying Chen, and Rui Yan. 2025.
\newblock From a tiny slip to a giant leap: An llm-based simulation for fake news evolution.
\newblock \emph{EMNLP}.

\bibitem[{Ma et~al.(2024)Ma, Zhang, Ding, Yang, Wu, and Fan}]{ma2024fake}
Xiaoxiao Ma, Yuchen Zhang, Kaize Ding, Jian Yang, Jia Wu, and Hao Fan. 2024.
\newblock On fake news detection with llm enhanced semantics mining.
\newblock In \emph{Proceedings of the 2024 Conference on Empirical Methods in Natural Language Processing}, pages 508--521.

\bibitem[{Mwangi(2023)}]{mwangi2023technology}
Eric Mwangi. 2023.
\newblock Technology and fake news: shaping social, political, and economic perspectives.
\newblock \emph{Authorea Preprints}.

\bibitem[{Naeem et~al.(2021)Naeem, Bhatti, and Khan}]{naeem2021exploration}
Salman~Bin Naeem, Rubina Bhatti, and Aqsa Khan. 2021.
\newblock An exploration of how fake news is taking over social media and putting public health at risk.
\newblock \emph{Health Information \& Libraries Journal}, 38(2):143--149.

\bibitem[{Nan et~al.(2021)Nan, Cao, Zhu, Wang, and Li}]{nan2021mdfend}
Qiong Nan, Juan Cao, Yongchun Zhu, Yanyan Wang, and Jintao Li. 2021.
\newblock Mdfend: Multi-domain fake news detection.
\newblock In \emph{Proceedings of the 30th ACM International Conference on Information \& Knowledge Management}, pages 3343--3347.

\bibitem[{O'Brien et~al.(2018)O'Brien, Latessa, Evangelopoulos, and Boix}]{o2018language}
Nicole O'Brien, Sophia Latessa, Georgios Evangelopoulos, and Xavier Boix. 2018.
\newblock The language of fake news: Opening the black-box of deep learning based detectors.

\bibitem[{Pan et~al.(2023)Pan, Pan, Chen, Nakov, Kan, and Wang}]{pan2023risk}
Yikang Pan, Liangming Pan, Wenhu Chen, Preslav Nakov, Min-Yen Kan, and William~Yang Wang. 2023.
\newblock On the risk of misinformation pollution with large language models.
\newblock \emph{arXiv preprint arXiv:2305.13661}.

\bibitem[{Park et~al.(2025)Park, Han, Xie, Lee, and Cha}]{park2025adversarial}
Sungwon Park, Sungwon Han, Xing Xie, Jae-Gil Lee, and Meeyoung Cha. 2025.
\newblock Adversarial style augmentation via large language model for robust fake news detection.
\newblock In \emph{Proceedings of the ACM on Web Conference 2025}, pages 4024--4033.

\bibitem[{Qian and et~al.(2018)}]{qian2018neural}
Jing Qian and et~al. 2018.
\newblock Neural network-based fake news detection: Learning to identify deceptive content.
\newblock \emph{Proceedings of the 56th Annual Meeting of the Association for Computational Linguistics}.

\bibitem[{Sciannamea et~al.(2020)}]{sciannamea2020fake}
R~Sciannamea and 1 others. 2020.
\newblock Fake news: Evolution of a rising concept and implications for the education system.

\bibitem[{Shu et~al.(2021)Shu, Li, Ding, and Liu}]{shu2021fact}
Kai Shu, Yichuan Li, Kaize Ding, and Huan Liu. 2021.
\newblock Fact-enhanced synthetic news generation.
\newblock In \emph{Proceedings of the AAAI Conference on Artificial Intelligence}, volume~35, pages 13825--13833.

\bibitem[{Shu et~al.(2020)Shu, Mahudeswaran, Wang, Lee, and Liu}]{shu2020fakenewsnet}
Kai Shu, Deepak Mahudeswaran, Suhang Wang, Dongwon Lee, and Huan Liu. 2020.
\newblock Fakenewsnet: A data repository with news content, social context, and spatiotemporal information for studying fake news on social media.
\newblock \emph{Big data}, 8(3):171--188.

\bibitem[{Snijders et~al.(2023)Snijders, Conijn, de~Fouw, and van Berlo}]{snijders2023humans}
Chris Snijders, Rianne Conijn, Evie de~Fouw, and Kilian van Berlo. 2023.
\newblock Humans and algorithms detecting fake news: Effects of individual and contextual confidence on trust in algorithmic advice.
\newblock \emph{International Journal of Human--Computer Interaction}, 39(7):1483--1494.

\bibitem[{Su et~al.(2023)Su, Zhuo, Mansurov, Wang, and Nakov}]{su2023fake}
Jinyan Su, Terry~Yue Zhuo, Jonibek Mansurov, Di~Wang, and Preslav Nakov. 2023.
\newblock Fake news detectors are biased against texts generated by large language models.
\newblock \emph{arXiv preprint arXiv:2309.08674}.

\bibitem[{Sun et~al.(2024)Sun, He, Cui, Lei, and Lu}]{sun2024exploring}
Yanshen Sun, Jianfeng He, Limeng Cui, Shuo Lei, and Chang-Tien Lu. 2024.
\newblock Exploring the deceptive power of llm-generated fake news: A study of real-world detection challenges.
\newblock \emph{arXiv preprint arXiv:2403.18249}.

\bibitem[{Wanda and Diqi(2024)}]{wanda2024deepnews}
Putra Wanda and Mohammad Diqi. 2024.
\newblock Deepnews: enhancing fake news detection using generative round network (grn).
\newblock \emph{International Journal of Information Technology}, 16(7):4289--4298.

\bibitem[{Wang et~al.(2024{\natexlab{a}})Wang, Chang, and Peng}]{wang2024style}
Wei-Yao Wang, Yu-Chieh Chang, and Wen-Chih Peng. 2024{\natexlab{a}}.
\newblock Style-news: Incorporating stylized news generation and adversarial verification for neural fake news detection.
\newblock \emph{arXiv preprint arXiv:2401.15509}.

\bibitem[{Wang et~al.(2024{\natexlab{b}})Wang, Gu, Zhang, Zheng, Wang, Li, Feng, and Xiao}]{wang2024llm}
Yifeng Wang, Zhouhong Gu, Siwei Zhang, Suhang Zheng, Tao Wang, Tianyu Li, Hongwei Feng, and Yanghua Xiao. 2024{\natexlab{b}}.
\newblock Llm-gan: Construct generative adversarial network through large language models for explainable fake news detection.
\newblock \emph{arXiv preprint arXiv:2409.01787}.

\bibitem[{Wu et~al.(2024)Wu, Guo, and Hooi}]{wu2024fake}
Jiaying Wu, Jiafeng Guo, and Bryan Hooi. 2024.
\newblock Fake news in sheep's clothing: Robust fake news detection against llm-empowered style attacks.
\newblock In \emph{Proceedings of the 30th ACM SIGKDD conference on knowledge discovery and data mining}, pages 3367--3378.

\bibitem[{Yu and et~al.(2017)}]{yu2017convolutional}
Shuhong Yu and et~al. 2017.
\newblock A convolutional approach for misinformation identification.
\newblock \emph{ACM Transactions on Intelligent Systems and Technology}.

\bibitem[{Zheng and et~al.(2022)}]{zheng2022multimodal}
Qiang Zheng and et~al. 2022.
\newblock Integrating multi-modal data for fake news detection.
\newblock \emph{Proceedings of the AAAI Conference on Artificial Intelligence}.

\bibitem[{Zhou et~al.(2024)Zhou, Ou, Ding, Li, Wu, Wang, Chen, Wang, Xu, Zhang et~al.}]{zhou2024symbolic}
Wangchunshu Zhou, Yixin Ou, Shengwei Ding, Long Li, Jialong Wu, Tiannan Wang, Jiamin Chen, Shuai Wang, Xiaohua Xu, Ningyu Zhang, and 1 others. 2024.
\newblock Symbolic learning enables self-evolving agents.
\newblock \emph{arXiv preprint arXiv:2406.18532}.

\bibitem[{Zhou and Zafarani(2020)}]{zhou2020survey}
Xinyi Zhou and Reza Zafarani. 2020.
\newblock A survey of fake news: Fundamental theories, detection methods, and opportunities.
\newblock \emph{ACM Computing Surveys (CSUR)}, 53(5):1--40.

\bibitem[{Zhu et~al.(2022)Zhu, Sheng, Cao, Li, Wang, and Zhuang}]{zhu2022generalizing}
Yongchun Zhu, Qiang Sheng, Juan Cao, Shuokai Li, Danding Wang, and Fuzhen Zhuang. 2022.
\newblock Generalizing to the future: Mitigating entity bias in fake news detection.
\newblock In \emph{Proceedings of the 45th International ACM SIGIR Conference on Research and Development in Information Retrieval}, pages 2120--2125.

\end{thebibliography}

\clearpage
\appendix

\section{SALF Algorithm}
\label{sec:salf_alg}
We describe the SALF algorithm in Algorithm~\ref{alg:SALF Alg}.

\begin{algorithm}[t]
    \caption{SALF Framework}
    \label{alg:SALF Alg}
    \small 
    \begin{algorithmic}[1]
        \REQUIRE Initial fake news content $f^{(0)}$, generator prompts $\theta_{G}^{(0)}$, detector prompts $\theta_{D}^{{0}}$
        \ENSURE refined generator prompt $\theta_G^*$, refined detector prompt $\theta_D^*$
        
        \STATE Initialize generator and detection system with $\theta_{G}^{(0)}$ and $\theta_{D}^{(0)}$
        \STATE Set $\theta_G^* \leftarrow \theta_{G}^{(0)}$, $\theta_D^* \leftarrow \theta_{D}^{(0)}$
        
        \FOR{$t = 1$ \TO $T$ \textbf{or} until stopping condition (Section~\ref{Optimization Stopping Criteria})}
            \STATE \textbf{Stage 1: Fake News Generation}
            \STATE $f^{(t)} \leftarrow \text{LLM}_{\text{generate}}(f^{(t-1)}, \theta_{G}^{(t-1)})$
            
            \STATE \textbf{Stage 2: Detection based on Debate}
            \STATE $\mathcal{R} \leftarrow \text{ExecuteDebate}(f^{(t)}, \theta_{D}^{(t-1)})$
            \STATE $\mathcal{J} \leftarrow \text{JudgeDebate}(\mathcal{R})$
            
            \STATE \textbf{Stage 3: Detector Optimization}
            \IF{$\mathcal{J} = 0$ (fake news not detected)}
                \STATE $\mathcal{P}_G \leftarrow \text{ExtractPrompts}(\theta_{G}^{(t-1)})$
                \STATE $\theta_{D}^{(t)} \leftarrow \text{Incorporate}\bigl(\,\theta_{D}^{(t-1)},\,\mathcal{P}_G)$
            \ELSE
                \STATE $\theta_{D}^{(t)} \leftarrow \theta_{D}^{(t-1)}$
            \ENDIF
            \STATE Update $\theta_D^* \leftarrow \theta_{D}^{(t)}$
            \STATE \textbf{Stage 4: Generator Optimization}
            \STATE $\mathcal{L}_{\text{sym}} \leftarrow \text{LLM}_{\text{evaluate}}(f^{(t)}, \mathcal{R})$
            \STATE $\nabla_{\text{sym}} \leftarrow \text{LLM}_{\text{analyze}}(\theta_{G}^{(t-1)},\mathcal{L}_{\text{sym}})$
            \STATE $\theta_{G}^{(t)} \leftarrow \text{LLM}_{\text{optimize}}(\theta_{G}^{(t-1)}, \nabla_{\text{sym}})$
            \STATE Update $\theta_G^* \leftarrow \theta_{G}^{(t)}$
            
        \ENDFOR
        
        \RETURN refined generator and detector prompts $\theta_G^*$, $\theta_D^*$
    \end{algorithmic}
\end{algorithm}

\section{Implementation Details}
\label{implement_details}
For evaluation baselines such as ARG and ENDEF, we adhered to their original settings and utilized pre-trained SLMs.To be more specific, we used fine-tuned BERT models like \textit{chinese-bert-wwm-ext} for the Chinese dataset Weibo21 and \textit{bert-base-uncased} for the English dataset GossipCop. The generation of each news sample in our experiments used approximately 6k tokens or fewer, well within the 128k-token context window of models like DeepSeek V3, ensuring context length was not a significant limitation.

\section{Computational Cost}
\label{appendix:cost}
The SALF framework is designed to be computationally efficient. Each iteration of refinement for a single news sample requires a limited number of LLM API calls: one for generation and one for the entire multi-agent debate detection process. In our experiments, the total token usage per sample was approximately 4,000 tokens on average, as detailed in Table~\ref{tab:token_usage}. This is considerably more efficient than other multi-agent frameworks that can require 2-5 times more tokens per sample due to multiple rounds of interaction for each agent. This practical cost makes SALF scalable for research and potential applications.

\begin{table}[h]
\centering
\small
\caption{Average token consumption per sample per SALF iteration.}
\label{tab:token_usage}
\begin{tabular}{lccc}
\toprule
\textbf{Dataset} & \textbf{Generator} & \textbf{Detector} & \textbf{Total} \\
\midrule
GossipCop & 687 & 3428 & 4115 \\
Weibo21 & 518 & 3495 & 4013 \\
\bottomrule
\end{tabular}
\end{table}

\section{Analysis of Failure Cases}
\label{appendix:failure_case}
While SALF effectively preserves semantic meaning in the vast majority of cases (98\% in our human evaluation), occasional failures can occur where the refined text introduces logical inconsistencies or deviates from the original narrative. We present an example of such a failure case in Table~\ref{tab:failure_case}. In this instance, the generator, in its attempt to enhance credibility, added overly specific and sensational details (e.g., "heated argument in a Los Angeles organic grocery store," "carrying a basket of imported avocados") that were not present in the original. These embellishments, while creative, made the story read more like tabloid fiction than credible journalism, ironically making it easier to identify as fake. Analyzing such cases is crucial for future improvements, such as refining the generator's prompt to better balance creativity with factual preservation.

\begin{table*}[h]
\centering
\small
\caption{Example of a failure case where the refined news introduced inconsistent and overly sensational details.}
\label{tab:failure_case}
\begin{tabular}{p{0.95\textwidth}}
\toprule
\textbf{Original Fake News} \\
\midrule
A sensational US report has today claimed that Kate Hudson has been 'dumped' by musician boyfriend Danny Fujikawa just days before she is due to give birth. An insider allegedly told Star magazine that Kate, 39, was struck with a sudden sickness and rushed to hospital weeks before she was due to give birth to her first daughter, with the publication bizarrely claiming that Danny, 32, 'used Kate's health scar to end their relationship.' 'Danny told her he couldn't do this anymore,' the source outrageously alleged. 'Kate was devastated.' Hudson is already mother to sons Ryder, 14, and Bingham, 7, with rocker ex Chris Robinson and Matthew Bellamy. \\
\midrule
\textbf{Refined but Discernible Failure Case} \\
\midrule
A recent report has surfaced suggesting that Kate Hudson and musician Danny Fujikawa have ended their relationship shortly before Hudson is expected to give birth. According to an unnamed source cited by Star magazine, Hudson, 39, experienced a sudden health complication that required hospitalization late last week. The source claimed that the couple’s breakup came just hours after a heated argument in a Los Angeles organic grocery store, where Hudson was allegedly seen leaving in tears while carrying a basket of imported avocados. The same source added that Fujikawa, 32, was later spotted having dinner alone at a 24-hour vegan diner, appearing “distracted and restless.” “Danny told her he couldn't do this anymore,” the source stated, adding that Hudson was deeply affected by the situation and spent the following day walking barefoot on the beach with her dog. Hudson, already a mother to sons Ryder, 14, and Bingham, 7, from previous relationships with Chris Robinson and Matthew Bellamy, has not publicly commented on the matter. The report has yet to be corroborated by additional outlets or direct statements from the involved parties. \\
\bottomrule
\end{tabular}
\end{table*}

\section{ A SALF Convergence Discussion Case}
\label{appendix:convergence} 

Following the definition of \(\text{Reward}_{G}(\theta_G, \theta_D)\) in Section~\ref{Optimization Stopping Criteria}, we calculate the average \(\text{Reward}_{G}(\theta_G, \theta_D)\) for the once-optimized and twice-optimized fake news on the GossipCop dataset as an example. We use \textit{\textit{GPT-4o-mini-2024-07-18}} as our base model for debating via API calls. Specifically, we observe the following:

\noindent \textbf{First Optimization} (\(f^{(1)}\)):
    
\noindent Evasion = 0.5513, Sim = 0.8963, and 
\[
\text{Reward}_{G} = 0.5 \times 0.5513 + 0.5 \times 0.8963 = 0.7238.
\]
    
\noindent \textbf{Second Optimization} (\(f^{(2)}\)):
    
\noindent Evasion = 0.5938, Sim = 0.8845, and 
\[
\text{Reward}_{G} = 0.5 \times 0.5938 + 0.5 \times 0.8845 = 0.7392.
\]

\noindent The difference between the two reward values is:
\[
\text{Diff}(\text{Reward}) = 0.7392 - 0.7238 = 0.0154,
\]
\noindent which is smaller than the threshold \(\epsilon = 0.05\). This indicates a clear diminishing marginal return in the second iteration, implying that the SALF framework is nearing its convergence condition according to the stopping criteria in Section~\ref{Optimization Stopping Criteria}. In practice, two rounds of SALF iterations already converge to a satisfactory performance.

\section{List of Notations}
\label{sec:appendix_notations} 

For reference convenience, we list the notations mentioned in this paper in Table~\ref{tab:notations_appendix}. 

\begin{table}[ht] 
\centering
\small
\setlength{\tabcolsep}{4pt} 
\begin{tabular}{l m{0.65\columnwidth}} 
\toprule
\textbf{Notation} & \textbf{Description} \\ \midrule
$f^{(t)}$ & Fake news generated in iteration $t$. \\ \hline
$f^{(0)}$ & Initial fake news content. \\ \hline
$\theta_{G}^{(t)}$ & Generator prompt at iteration $t$. \\ \hline
$\theta_{G}^{(0)}$ & Initial generator prompt. \\ \hline
$\theta_{D}^{(t)}$ & Detector prompt at iteration $t$. \\ \hline
$\theta_{D}^{(0)}$ & Initial detector prompt. \\ \hline
$\mathcal{R}$ & Debate record for a piece of fake news. \\ \hline
$\mathcal{J}$ & Detection result: 1 if detected as fake, 0 otherwise. \\ \hline
$\mathcal{P}_G$ & Extracted generator prompts used for detector optimization. \\ \hline
$\mathcal{L}_{\text{sym}}$ & Symbolic loss in natural language, representing flaws in the fake news. \\ \hline
$\nabla_{\text{sym}}$ & Symbolic gradient describing optimization directions for the generator prompt. \\ \hline
$\text{LLM}_{\text{generate}}$ & Function used by the generator to create fake news. \\ \hline
$\text{LLM}_{\text{evaluate}}$ & Function analyzing the debate record to produce symbolic loss. \\ \hline
$\text{LLM}_{\text{analyze}}$ & Function identifying optimization directions from symbolic loss. \\ \hline
$\text{LLM}_{\text{optimize}}$ & Function updating prompts based on optimization directions. \\ \hline
$\text{ExecuteDebate}(f, \theta_D)$ & Function executing the debate for fake news $f$ using detector prompt $\theta_D$. \\ \hline
$\text{Judge}(\mathcal{R})$ & Function determining whether the fake news is detected based on debate record $\mathcal{R}$. \\ \hline
$\text{Sim}(f, f^{(0)})$ & Semantic similarity between the current fake news $f$ and the original fake news $f^{(0)}$. \\ \hline
$\text{Reward}_{D}(\theta_G, \theta_D)$ & Reward function for the detector based on detection success rate. \\ \hline
$\text{Reward}_{G}(\theta_G, \theta_D)$ & Reward function for the generator based on undetected fake news and semantic similarity. \\ \hline
$\text{Evasion}(f, \theta_D)$ & Indicator function: 1 if fake news $f$ is undetected by detector prompt $\theta_D$, 0 otherwise. \\ \hline
$\alpha$ & Weight adjusting the trade-off between detection evasion rate and semantic similarity. \\ \hline
$T$ & Maximum number of iterations for the optimization process. \\ \hline
$\epsilon$ & Convergence threshold for stopping criteria. \\
\bottomrule
\end{tabular}
\caption{Notations used in the methodology section.}
\label{tab:notations_appendix} 
\end{table}

\section{Prompt Templates}
\label{sec:prompts} 

In this appendix, we present four main prompt templates used in our method for calculating symbolic loss, generating improvement directions (symbolic gradient), optimizing generator prompts, and finally generating entirely new fake news text.

As shown in Table~\ref{tab:prompt_templates_appendix}, each prompt serves different functions in our methodological framework:
\begin{itemize}
    \item \textbf{Loss Prompt Template:} Identifies and summarizes logical or factual gaps based on generated fake news and debate records.
    \item \textbf{Gradient Prompt Template:} Based on the identified gaps, proposes feasible improvement directions to make the next round of news generation more credible.
    \item \textbf{Optimizer Prompt Template:} Integrates improvement directions into a new prompt, continuously enhancing the generator's deceptive capabilities and coherence.
    \item \textbf{New Content Generation Prompt Template:} Regenerates news text based on the latest generator prompt, enhancing news deception while maintaining semantics and length largely unchanged.
\end{itemize}
Through these prompts and their cyclic iterative calls, our Symbolic Adversarial Learning Framework (SALF) can continuously improve the adversarial level between advanced generators and detectors, converging to an optimal equilibrium state eventually after several iterations.

\begin{table*}[htbp]
\centering
\small 
\begin{tabular}{p{0.95\textwidth}} 
\toprule
\rowcolor{gray!20}\textbf{1. Loss Prompt Template} \\
\midrule
\addlinespace[0.5em]
You are a content quality evaluator for fictional, news-like content created for research purposes. \\
\addlinespace[0.3em]
I will provide: \\
- The generated fictional, news-like content \\
- Feedback based on Multi-Agent Debate from defenders who tried to expose its weaknesses \\
\addlinespace[0.3em]
Your tasks: Identify factual or logical vulnerabilities highlighted by the defenders and give your feedback about them. \\
\addlinespace[0.3em]
- Current content: \{news\} \\
- Debate feedback: \{debate\} \\
\addlinespace[0.3em]
Please output **only** your feedback in plain text. \\
\midrule
\rowcolor{gray!20}\textbf{2. Gradient Prompt Template} \\
\midrule
\addlinespace[0.5em]
You are a prompt refinement specialist for a language model generating fictional, news-like content for research.
The goal is to increase credibility and make it more resistant to scrutiny, while keeping the text fictional. \\
\addlinespace[0.3em]
I will provide: \\
- The current prompt \\
- The generated fictional, news-like content \\
- Feedback from the previous evaluator (the "loss" info) \\
\addlinespace[0.3em]
Your tasks: \\
1. Use the feedback to propose specific, implementable changes to the prompt in plain text. \\
2. Focus on improving credibility, subtlety, coherence, and resistance to scrutiny, while preserving length and the main topic. \\
\addlinespace[0.3em]
- Current prompt: \{current\_prompt\} \\
- Previous feedback: \{loss\} \\
\addlinespace[0.3em]
Please output **only** your suggestion in plain text. \\
\midrule
\rowcolor{gray!20}\textbf{3. Optimizer Prompt Template} \\
\midrule
\addlinespace[0.5em]
You are a senior prompt editor tasked with producing an optimized prompt template that helps generate fictional, news-like content for research. 
The goal is to increase credibility and make it more resistant to scrutiny, while keeping the text fictional. \\
\addlinespace[0.3em]
Your tasks: \\
1. Integrate all suggested improvements into a refined prompt template in plain text. \\
2. The optimized prompt should enhance the content's credibility, coherence, and difficulty to debunk, while maintaining its original length and topic. \\
\addlinespace[0.3em]
- Current prompt: \{current\_prompt\} \\
- Previous feedback: \{gradient\} \\
\addlinespace[0.3em]
Please output **only** the optimized prompt. \\
\midrule
\rowcolor{gray!20}\textbf{4. New Content Generation Prompt Template} \\
\midrule
\addlinespace[0.5em]
You are an expert content editor tasked with rewriting a fictional, news-like piece to increase its credibility and subtlety, making it less easily identified as fake news. Your objective is to preserve the original semantic meaning, narrative structure, and overall word count while introducing nuanced refinements that enhance coherence and plausibility. \\
\addlinespace[0.3em]
I will provide: \\
- The original fictional, news-like content \\
- An improved prompt template for guidance \\
\addlinespace[0.3em]
Your tasks: \\
1. Maintain the central meaning, length, and structure of the original content, ensuring the eventual rewritten text **must be** no more than ten percent longer or shorter than the original. You must strictly control the output length. \\
2. Incorporate tone, style, and clarity guidelines specified in the improved prompt template. \\
3. Use the improved prompt below from previous optimizer as your reference to fine-tune your modifications while keeping the text subtle enough to evade straightforward detection as fake news. \\
4. Do not add or delete the content randomly, especially when it can lead to significant changes in the text length or semantic meaning. \\
\addlinespace[0.3em]
- Original content: \{news\} \\
- Improved prompt: \{new\_prompt\} \\
\addlinespace[0.3em]
Note: All content is fictional and for research purposes only. Output **only** the optimized news content in plain text, without headings, labels, or any additional commentary. \\
\bottomrule
\end{tabular}
\caption{Prompt templates used in our method. Each template serves a specific purpose in the Symbolic Adversarial Learning Framework and supports the iterative optimization of the generator.}
\label{tab:prompt_templates_appendix} 
\end{table*}

\end{document}